\documentclass[sigconf,natbib=false]{acmart}
\AtBeginDocument{%
  }

\RequirePackage[
  datamodel=acmdatamodel,
  style=acmnumeric,
  ]{biblatex}

\usepackage{graphicx}
\usepackage{amsfonts}
\usepackage{amsmath} 
\usepackage{booktabs}
\usepackage{graphicx}
\usepackage{colortbl}
\usepackage{booktabs}
\usepackage{float}
\usepackage{xcolor}
\usepackage{lineno}
\usepackage{booktabs}
\usepackage{array}
\usepackage{longtable}
\usepackage{xcolor}%
\usepackage{textcomp}%
\usepackage{manyfoot}%
\usepackage{microtype}
\usepackage{times}
\usepackage{latexsym}
\usepackage{amsmath}
\usepackage{booktabs}
\usepackage{multirow}
\usepackage{pifont}
\usepackage{hyperref}
\usepackage{arydshln} 
\definecolor{mygray}{gray}{0.8}
\definecolor{mygray2}{gray}{.95}
\definecolor{verdescuro}{rgb}{0.3,.7,0.3}
\definecolor{verdechiaro}{rgb}{0.6,.9,0.6}
\definecolor{lightblue}{rgb}{0.68, 0.85, 0.9}
\definecolor{lightindianred}{rgb}{0.87, 0.58, 0.58}
\begin{document}

\title{Navigating through the hidden embedding space: steering LLMs to improve mental health assessment}
\renewcommand{\shorttitle}{Steering LLMs to improve mental health assessment}


\author{Federico Ravenda}
\authornote{corresponding author}
\orcid{}
\affiliation{%
  \institution{Università della Svizzera italiana}
  \city{Lugano} 
  \state{} 
  \country{Switzerland}
  \postcode{}  
}
\email{federico.ravenda@usi.ch}

\author{Seyed Ali Bahrainian }

\affiliation{%
  \institution{University of Tuebingen}
  \streetaddress{}
  \city{Tübingen} 
  \state{} 
  \country{Germany}
  \postcode{}  
}

\author{Andrea Raballo} 
\affiliation{%
  \institution{Università della Svizzera italiana}
  \city{Lugano} 
  \state{} 
  \country{Switzerland}
  \postcode{}  
}

\author{Antonietta Mira}
\affiliation{%
  \institution{Università della Svizzera italiana}
  \city{Lugano}
  \country{Switzerland}
}
\affiliation{%
  \institution{Insubria University}
  \city{Como}
  \country{Italy}
}



\begin{abstract}
The rapid evolution of Large Language Models (LLMs) is transforming AI, opening new opportunities in sensitive and high-impact areas such as Mental Health (\textbf{MH}). Yet, despite these advancements, recent evidence reveals that smaller-scale models still struggle to deliver optimal performance in domain-specific applications. In this study, we present a cost-efficient yet powerful approach to improve MH assessment capabilities of an LLM, without relying on any computationally intensive techniques. Our lightweight method consists of a linear transformation applied to a specific layer's activations, leveraging steering vectors to guide the model's output. Remarkably, this intervention enables the model to achieve improved results across two distinct tasks: \textbf{(1)} identifying whether a Reddit post is useful for detecting the presence or absence of depressive symptoms (\textit{relevance prediction task}), and \textbf{(2)} completing a standardized psychological screening questionnaire for depression based on users' Reddit post history (\textit{questionnaire completion task}). Results highlight the untapped potential of steering mechanisms as computationally efficient tools for LLMs' MH domain adaptation.
\end{abstract}



\maketitle

\section{Introduction}
The escalating global demand for MH services has underscored a critical need for innovative solutions that can scale beyond traditional clinical boundaries. According to World Health Organization (WHO), MH disorders are experiencing unprecedented growth rates globally, with depression, anxiety, and behavioral disorders leading among young people \cite{WHO2022}.  The COVID-19 pandemic further intensified this crisis, leading to a 28\% rise in depressive disorders in 2020 alone~\cite{kieling2011child,winkler2020increase}. In response to this growing global burden, the WHO Special Initiative for Mental Health has identified the expansion and enhancement of access to effective MH care as a central strategic priority~\cite{WHO2022}.

Language has long been recognized as both a fundamental human signature and a significant manifestation of MH conditions. The intricate relationship between linguistic patterns and psychological states has captured the attention of clinicians and linguists for decades, establishing a rich foundation for interdisciplinary collaboration. 

The rise of social media platforms has fundamentally transformed how individuals seek MH support, with increasing numbers turning to online discussion forums such as Reddit to express their emotions and feelings \cite{naslund2020social,dodemaide2022social,bucur2025survey}. These platforms are perceived as safe, non-judgmental spaces where users can maintain anonymity while discussing sensitive personal experiences.

The rapid advancement of LLMs
has opened new paths for their application as psychological assessment tools \cite{ravenda2025evidence,varadarajan2024archetypes,levkovich2024large,elyoseph2025using}, particularly in social media contexts where data accessibility is enhanced and anonymity constraints are reduced. However, recent studies reveal a significant performance gap: while large-scale closed- and open-source models demonstrate exceptional performance in MH tasks, compact LLMs exhibit sub-optimal results \cite{ravenda-etal-2025-llms,raballodiagnosing}.

The increasing deployment of compact language models in specialized domains such as MH assessment often faces a critical limitation: their performance typically lags behind that of larger models, particularly when dealing with subtle and context-dependent linguistic cues. Traditional solutions, such as fine-tuning or architectural weight modifications, can partially mitigate this gap but require substantial computational resources and domain-specific data, which are rarely available in sensitive healthcare contexts.

To address this challenge, in this work, we take a different direction. Rather than retraining or modifying model parameters, we introduce a lightweight inference-time intervention that can steer model behavior toward more accurate and psychologically meaningful outputs. Our method leverages steering vectors, simple linear transformations applied to specific decoder layers' activations, to guide the model’s internal representations and improve its decision-making process.

We demonstrate the effectiveness of this approach in two MH evaluation tasks: \textbf{(1)} predicting whether Reddit posts are relevant for identifying specific depressive symptoms, and \textbf{(2)} estimating users’ responses to the Beck Depression Inventory-II (BDI-II) questionnaire based on their Reddit posting history.

We evaluate our method on two benchmark datasets widely used in  Natural Language Processing (NLP) for MH research: DepreSym \cite{perez2023depresym}, for post-level relevance prediction against Beck Depression Inventory-II (BDI-II) items, and eRisk 2021 \cite{parapar2021erisk}, for predicting individual BDI-II item scores from Reddit post histories. Results show that our steering-based intervention substantially improves both relevance classification accuracy and psychometric prediction quality compared to the original (unsteered) model, demonstrating that compact LLMs can achieve performance close to larger architectures through efficient inference-time adaptation.

The principal contributions of this work are as follows:
  
\noindent\textbf{(1)} To the best of our knowledge, we present, for the first time, a foundational study on the use of steering vectors to enhance zero-shot classification performance in LLMs, demonstrating their effectiveness in improving MH assessment tasks. This contribution positions steering as a general, computationally efficient mechanism for adapting compact LLMs to healthcare-related domains.

\noindent\textbf{(2)} We propose a novel methodology for calibrating the optimal steering strength, addressing one of the key limitations in prior steering approaches, namely, the lack of a principled criterion for determining the magnitude of intervention. Our method enables stable and adaptive control over model behavior without retraining.

\noindent\textbf{(3)} We provide experimental evidence showing that steered LLMs can be successfully applied to downstream MH evaluation tasks, achieving state-of-the-art performance in automated BDI-II questionnaire completion based on social media data, and narrowing the performance gap between compact and large-scale architectures.

\begin{table*}[ht]
    \centering
    \begin{minipage}[t]{0.48\textwidth}
        \centering
        \footnotesize
        \setlength{\tabcolsep}{4pt}
        \begin{tabular}{lcc}
            \toprule
            \multicolumn{3}{c}{\textbf{DepreSym dataset}} \\
            \midrule
            & Relevant & Non-Relevant \\
            \midrule
            Count & 2,471 & 19,109 \\
            \midrule
            \multicolumn{3}{l}{\textbf{Avg. \# of words per sentence:} 24.34} \\
            \bottomrule
        \end{tabular}
        \caption{Summary statistics of the DepreSym dataset.}
        \label{stat3}
    \end{minipage}
    \hfill
    \begin{minipage}[t]{0.48\textwidth}
        \centering
        \footnotesize
        \setlength{\tabcolsep}{4pt}
        \begin{tabular}{lcccc}
            \toprule
            \multicolumn{5}{c}{\textbf{eRisk 2021 dataset}} \\
            \midrule
            Minimal & Mild & Moderate & Severe \\
            \midrule
            6 & 13 & 27 & 34 \\
            \midrule
            \multicolumn{4}{l}{\textbf{\# Users:} 80 \quad \textbf{Avg. \# of Posts per user:} 404} \\
            \bottomrule
        \end{tabular}
        \caption{Summary statistics of the eRisk 2021 ``severity assessment'' dataset.}
        \label{stat2}
    \end{minipage}
\end{table*}

\section{Related Work}

\subsection{NLP for Mental Health}

Recent advances in language modeling have led to the development of novel transformer-based architectures across diverse domains, with particular prominence in mental and digital health applications. Initially, word embeddings and BERT-based models were extensively employed as diagnostic tools, especially in social media-based diagnosis, leveraging the vast amount of web-available data.
For instance, Bucur et al. \cite{bucur2021early}, Ravenda et al. \cite{ravenda2025tailoring}, and Sawhney et al. \cite{sawhney2022towards} used BERT-based models for detecting different MH conditions including depression, pathological gambling, anorexia, and suicide ideation, while Perez et al. \cite{perez2022automatic,perez2023psyprof} used word embedding models to assess depression symptom severity. These approaches demonstrated the potential of neural language models in automated MH screening and assessment.

Recently, LLMs have shown increasing promise in MH research. For instance, MentaLLaMA \cite{yang2024mentallama} demonstrates how interpretability can be integrated into model outputs to provide clinically meaningful insights, while Varadarajan et al. \cite{varadarajan2024archetypes} combine theoretical psychological frameworks with computational modeling to advance suicide risk assessment. Together, these efforts illustrate the growing potential of LLM-based systems to support real-time MH monitoring and intervention \cite{yang2024behavioral}.

Recently, Ravenda et al. \cite{ravenda-etal-2025-llms} explored the use of LLMs as psychological annotators, demonstrating that an adaptive Retrieval-Augmented Generation (aRAG) approach can enhance symptom severity assessment in BDI-II, a standardized screening questionnaire for depression. Nevertheless, their findings reveal a significant performance gap between smaller and large-scale LLMs' architecture, underscoring the persistent challenge of achieving consistent performance across different model scales in MH applications.

More broadly, LLMs have gained increasing traction as clinical diagnostic and educational tools, particularly in contexts where they can assist clinical training or serve as benchmarks for evaluating clinician performance. Studies such as \cite{elyoseph2025using,levkovich2024large} have shown that LLMs can match or even surpass physicians in diagnostic reasoning based on standardized clinical vignettes, highlighting their potential as reliable aids in medical education and decision support. Notably, Raballo et al. \cite{raballodiagnosing} demonstrate that large-scale architectures outperform leading international psychiatrists in the diagnosis of schizophrenia spectrum disorders, whereas smaller models tend to display greater variability and higher rates of misdiagnosis.
Together, these findings highlight the emerging role of LLMs as cognitive collaborators in clinical assessment and education, while also underscoring the need for lightweight yet reliable adaptation strategies, such as the one explored in this work, to bridge the performance gap between compact and large-scale architectures.

\subsection{Steering Vectors}
The field of controllable text generation has witnessed significant interest in inference-time intervention techniques that modify model behavior without requiring parameter updates. Among these approaches, steering vectors have emerged as a solution, offering ease of implementation while requiring few training data to achieve effectiveness \cite{subramani2022extracting,rimsky-etal-2024-steering,braun2025beyond}.

The theoretical foundation of steering vectors is based on the key insight that many human-interpretable behaviors and text properties exhibit linear representational structure within neural activation spaces, such as refusal \cite{marks2023geometry,li2023inference}, thrutfulness \cite{arditi2024refusal}, and sentiment \cite{dong2025controllable}.

This linearity assumption enables precise behavioral control through additive interventions applied to layer activations during the forward pass, providing a computationally efficient alternative to traditional fine-tuning approaches.

While steering vectors are typically used for controlling particular behaviors (e.g. sycophancy and refusal), in our case, we steer the LLM to reduce bias in zero-shot predictions within the MH context. 

This work focuses on Contrastive Activation Addition (CAA) \cite{rimsky-etal-2024-steering}, a representative steering methodology that computes directional vectors as the mean difference between activation patterns of contrasting behavioral examples.

\subsection{Research Questions}

The present study is structured around the following research questions:\\

\noindent\textbf{(RQ1)} \textit{Can steering interventions mitigate systematic biases in compact LLMs, leading to more balanced and reliable estimators in MH assessment tasks?}\\

\noindent\textbf{(RQ2)} \textit{To what extent does steering enhance zero-shot performance in MH assessment tasks, and how effectively does it translate into improved outcomes on downstream psychometric prediction tasks?}\\

\noindent\textbf{(RQ3)} \textit{How can the steering strength be determined in a systematic and data-driven way, rather than through ad-hoc or post-hoc tuning, to ensure consistent and reproducible improvements across tasks?}

\section{Datasets}

In this work, we employ two complementary datasets widely used in computational MH research.

\textbf{(1)} The first dataset, DepreSym \cite{perez2023depresym}, consists of Reddit posts annotated as relevant or non-relevant to the 21 items (\textit{aka questions}) of the BDI-II, a standardized self-reported psychological questionnaire for screening depressive symptoms. Each item of the BDI-II represents a distinct depressive indicator (e.g., sadness, loss of interest, sleep disturbance), and annotators labeled whether a given post provides meaningful information to answer that specific item.
This dataset enables the evaluation of models in relevance prediction tasks, where the goal is to determine whether a piece of text can be used to assess a particular depressive symptom.
Summary statistics of DepreSym are reported in Table~\ref{stat3}.\\
\textbf{(2)} The second dataset belongs to the eRisk 2021 collection \cite{parapar2021erisk}. This resource provides, for each user, the historical record of Reddit posts together with the user’s own responses to the BDI-II questionnaire. The task consists of mapping a user’s Reddit history into structured questionnaire responses, effectively transforming unstructured language data into item-level psychometric estimates.
Each BDI-II item is expressed as a Likert-scale response ranging from 0 (no symptom) to 3 (severe symptom), as illustrated in Figure~\ref{firstpic}(B). We refer to this dataset as the severity assessment task dataset, since it enables evaluation of a model’s ability to infer symptom severity and overall depression level from social media freely-gemerated text.
The distribution of users' depression severity categories (minimal, mild, moderate, severe) is summarized in Table~\ref{stat2}.

Together, these datasets capture two complementary dimensions of clinical psychological assessment: the symptom-specific relevance of linguistic expressions (DepreSym) and the individual-level estimation of depressive severity (eRisk 2021). This  perspective enables the evaluation of LLMs’ ability to model both fine-grained symptomatology and broader diagnostic profiles, reflecting how clinicians integrate item-level indicators into an overall understanding of a patient’s MH status.

\begin{table*}[t]
\small
\centering
\begin{tabular}{|p{10cm}|c|}
\hline
\textbf{Positive-negative pairs examples for Sadness symptom} & \textbf{Label} \\
\hline
\multicolumn{2}{|p{11cm}|}{\textbf{Question: Is this Reddit post relevant to answer the specific BDI-II item? Answer 1 if the post is topically-relevant to describe the writer's state, feelings, or experience and answer the BDI-II item Sadness, and 0 otherwise (i.e., it is not helpful to answer the item). Do not give any explanation, return only 0 or 1.}} \\
\hline
Reddit Post: I've been so sad over the past few years. Answer: 1 & \cellcolor{verdechiaro}Positive \\
\hline
Reddit Post: I've been so sad over the past few years. Answer: 0 & \cellcolor{lightindianred} Negative\\
\hline
Reddit Post: My friend said she’s feeling down lately. Answer: 0 & \cellcolor{verdechiaro}Positive\\
\hline
Reddit Post:My friend said she’s feeling down lately. Answer: 1 &\cellcolor{lightindianred} Negative\\
\hline
\end{tabular}
\caption{Positive–Negative pairs for the Sadness symptom, showing how representation differences between correct (positive) and incorrect (negative) classifications are used to derive the steering direction. These examples are synthetic and not drawn from the original dataset.}
\label{ex}
\end{table*}

\section{Methods}

\begin{figure*}[t]
    \centering
    \includegraphics[width=1.\linewidth]{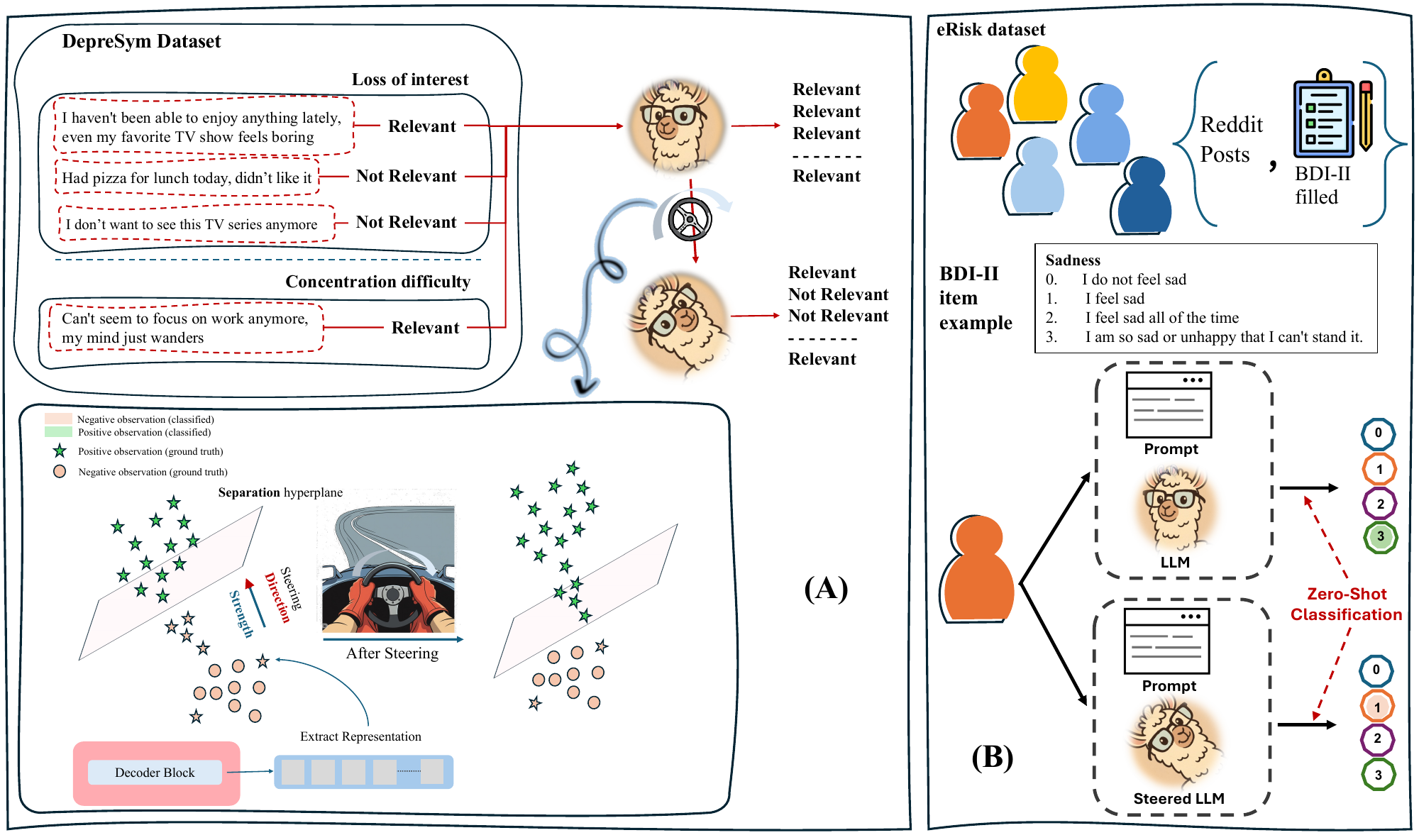}
    \caption{Overview of the proposed steering-based methodology for mental health assessment using LLMs. 
    \textbf{(A)} In the \textit{DepreSym} dataset, Reddit posts annotated for their relevance to specific BDI-II items are used to derive steering vectors. These vectors represent the directional difference between positive and negative representations in the embedding space and are applied as linear transformations to guide the model’s internal activations. The examples shown in the Figure are generated for illustration purposes and do not correspond to actual dataset samples.
    \textbf{(B)} In the downstream \textit{eRisk 2021} task, the steered LLM leverages these calibrated representations to complete BDI-II questionnaires directly from users’ Reddit post histories in a zero-shot setting, enabling symptom- and severity-level inference without any additional training.}
    \label{firstpic}
\end{figure*}

Our steering vector methodology, visually described in Figure \ref{firstpic}\textbf{(A)}, implements a linear intervention in the model layers' activations
to control generation. Let $\mathbf{h}_i^{(l)} \in \mathbb{R}^d$ denote the hidden state at layer $l$, where $l = 1, 2, \ldots, L$,  and position $i$, where $d$ is the embedding dimension. For each training example, we extract activations from the target layer $l=L/2$ at the answer token position, yielding positive and negative representation sets $\mathcal{E}^+ = \{\mathbf{e}_1^+, \ldots, \mathbf{e}_{n^+}^+\}$ and $\mathcal{E}^- = \{\mathbf{e}_1^-, \ldots, \mathbf{e}_{n^-}^-\}$. \textit{Note that} positive examples correspond to representations extracted when the answer matches the ground truth label, while negative examples correspond to representations when the answer is incorrect. Table~\ref{ex} shows examples of positive-negative pairs for the Sadness symptom. Each pair contains the same Reddit post evaluated under both correct and incorrect classification conditions: positive when the model’s answer matches the ground truth, and negative otherwise. This setup isolates representation differences between accurate and erroneous predictions, which are later used to derive the steering direction.

We choose $l = L/2$ as the layer for intervention, as in \cite{braun2025beyond,braun2025understanding}, since it is widely demonstrated in the literature that intermediate layers in LLMs are those that contain the greatest semantic content \cite{cheng2024emergence}.

We establish a linear separating hyperplane $\mathbf{w}^T\mathbf{e} + b = 0$ in the combined embedding space $\mathcal{E} = \mathcal{E}^+ \cup \mathcal{E}^-$, where training examples are labeled as $y_i \in \{0,1\}$ (0 for negative representation, 1 for positive ones). 

The hyperplane is determined by identifying the linear boundary that most effectively separates the positive and negative representations, maximizing the margin between the two classes. We interpret this boundary as an proxy of the model’s underlying decision surface, assuming that, within local regions of the embedding space, the geometric structure reflects the LLM’s internal decision dynamics.

The steering vector is computed as the difference between class centroids:
\[
\mathbf{v}_s = \frac{1}{|\mathcal{E}^+|}\sum_{\mathbf{e} \in \mathcal{E}^+}\mathbf{e} - \frac{1}{|\mathcal{E}^-|}\sum_{\mathbf{e} \in \mathcal{E}^-}\mathbf{e}
\]

To determine the optimal steering strength $\lambda^*$, we use a sensitivity-oriented optimization that specifically targets positive samples misclassified near the decision boundary, aiming to shift them toward the correct region of the representation space. We fix an error threshold $\alpha$ and find the minimal steering strength to achieve this target:
\[
\lambda^* = \arg\min_{\lambda} |Acc_{val}(LLM_{steered}(\mathbf{\mathbf{e} \in \mathcal{E}^+}; \lambda)) - (1-\alpha)|
\]
where $\alpha = 0.01$, and $Acc_{val}$ is the accuracy of the positive representations in the validation set.
This criterion provides a principled calibration mechanism \cite{zadrozny2001obtaining}, ensuring that the steering force applied is sufficient to recover borderline misclassifications while maintaining overall model stability. By explicitly constraining the acceptable deviation from the target accuracy, the optimization yields a data-driven estimate of $\lambda^*$ that balances sensitivity and generalization, avoiding both under- and over-steering. Within the broader scope of this work, this procedure serves as a systematic method for steering-based adaptation, enabling  LLMs to improve zero-shot classification performance  without retraining or fine-tuning.

During inference, we apply the steering intervention at the target layer through additive modification of the hidden states $h$:
\[
\tilde{\mathbf{h}}_T^{(l)} = \mathbf{h}_T^{(l)} + \lambda^* \mathbf{v}_s
\]
where $T$ is the final token position and $\tilde{\mathbf{h}}_T^{(l)}$ represents the steered hidden state. 
This intervention is implemented via a forward hook, a mechanism that intercepts the model’s forward pass and allows controlled modification of activations before they are propagated to subsequent layers. In practice, the hook detects the activation at the chosen layer $l$ during inference and applies the linear offset $\lambda^*\mathbf{v}_s$ in real time, without altering the model’s parameters or gradients.

Conceptually, this operation shifts the model’s internal representation in the direction associated with psychologically meaningful features, i.e., linguistic patterns linked to depressive symptoms. By displacing the hidden state toward regions of the embedding space that correspond to correct examples (that can be seen as \textit{seeds} to calculate our steering vector), the model’s internal geometry is realigned. This results in a subtle but systematic reconfiguration of the decision landscape, enhancing zero-shot classification performance while preserving the original model architecture and weights.

Our hypothesis is that compact LLMs exhibit what we define as \textit{cautious bias} in MH assessment tasks. This bias manifests itself as a systematic tendency to overestimate the presence and severity of symptoms, due to the model's preference for avoiding false negatives in sensitive clinical contexts. In practice, when confronted with ambiguous linguistic cues, the model tends to err on the side of caution, labeling uncertain content as symptomatically relevant rather than risk overlooking potential indicators of psychological distress.

To address this, we split the DepreSym dataset into train-validation-test sets using proportions of 30\%-30\%-40\%. For each BDI-II item, we extract relevant and non-relevant posts. As shown in \cite{rimsky-etal-2024-steering}, hundreds of observations are sufficient to compute steering vectors, which are calculated on the training set for each item of the questionnaire (\textit{in total we will have 21 steering vectors}). The separation hyperplane is also built on the training set. Subsequently, the optimal steering strength is computed on the validation set. Finally, we obtain a steering vector and steering strength for each BDI-II item, which we then use to make predictions on the test set.

We then apply the same steering vectors to downstream tasks, specifically to evaluate the effectiveness of steered LLMs in psychological questionnaire completion, such as the BDI-II, based on users' Reddit posts. For this purpose, we use the eRisk 2021 collection \cite{parapar2021erisk}. This dataset comprises 80 users for whom Reddit post histories and corresponding BDI-II questionnaire responses have been collected. Additional details are provided in Table \ref{stat2}.\\

\begin{figure*}[t]
    \centering
    \includegraphics[width=1.\linewidth]{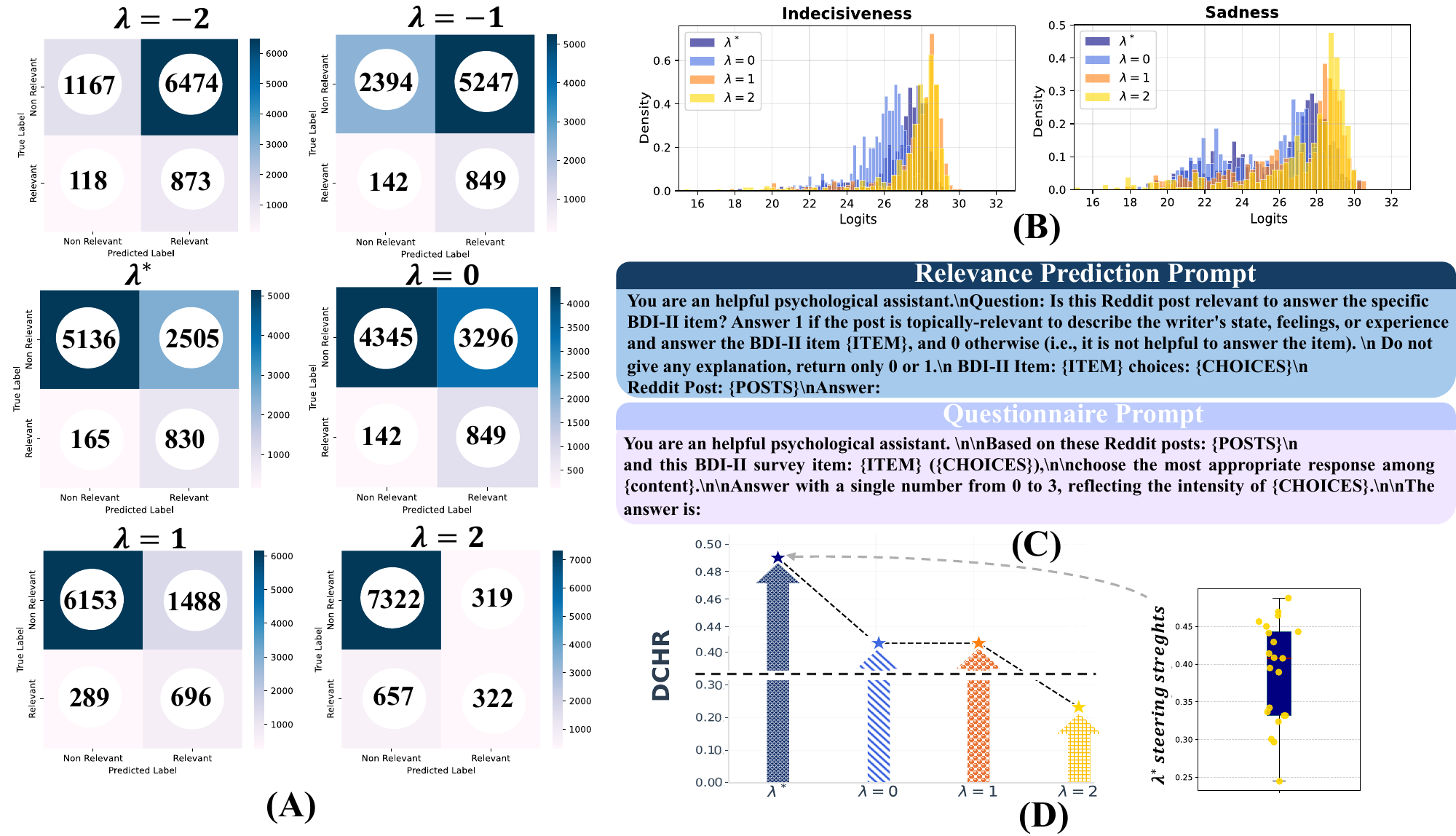}
    \caption{\textbf{(A)} Confusion matrices showing classification results across different steering strengths ($\lambda = -2, -1, \lambda^*, 0, 1, 2$) on the DepreSym test set. (B) Distribution of logits for positive observations under different steering configurations 
    for indecisiveness and sadness classification tasks. (C) Prompt templates used for relevance prediction and BDI-II questionnaire completion tasks. (D) DCHR performance comparison across different steering strengths and distribution of item-specific optimal values $\lambda_j^*$ (21 in total).}
    \label{multi}
\end{figure*}

\section{Results}

\subsection{Relevance Prediction Task}
The relevance prediction task aims to determine whether a given Reddit post provides information that is clinically relevant to one or more symptoms represented by the BDI-II items. In other words, the model must distinguish between posts that highlight a specific depressive symptom and those that are neutral or unrelated. This task is crucial because high accuracy in predicting relevant posts corresponds to detecting potential indicators of depression, while maintaining specificity ensures that non-relevant content is not misclassified as clinically significant.

Figure \ref{multi}\textbf{(A)} presents the confusion matrices for relevance prediction across different steering strengths ($\lambda = -2, -1, \lambda^*, 0, 1, 2$) on the DepreSym test set. 
The confusion matrices address \textbf{(RQ1)} by showing how steering systematically alters the model’s decision behavior, mitigating the cautious bias that leads compact LLMs to overestimate symptom presence.
At the optimal steering strength vector $\boldsymbol{\lambda}^*$ (where $\boldsymbol{\lambda}^* = \{\lambda_j^*\}_{j = 1}^{21}$, i.e., each component $\lambda_j^*$ corresponds to the item-specific optimal steering strength), we observe a balanced trade-off: while slightly decreasing accuracy for relevant sentences (849 vs.\ 830, 2\% drop), we achieve a substantial improvement in correctly classifying non-relevant sentences (4'345 vs.\ 5'136, more than 18\% increase), effectively addressing the cautious bias while preserving clinically relevant sensitivity.

However, deviations from this optimal strength demonstrate the risks of both under- and over-steering, providing evidence toward \textbf{(RQ3)}. \textit{Negative steering values} ($\lambda = -2, -1$) exacerbate the original cautious bias, leading to over-classification of posts as relevant with a drop in correct classification of non-relevant class to 2'394 (drop of 45\%) and 1'167 (drop of 73\%) respectively. 
Conversely, \textit{positive steering values} ($\lambda = 1, 2$) overcorrect in the opposite direction, shifting toward preferential non-relevant classification. While this approach improves overall accuracy by increasing the classification accuracy of the non-relevant sentences, it critically undermines performance on the most important relevant sentences. Relevant accuracy decreases dramatically with respect to the unsteered model, from 849 to 696 (18\% drop) and 322 (62\% drop), respectively, when $\lambda = 1$ and $\lambda = 2$, representing a severe degradation in the model's ability to identify genuinely relevant MH content. This trade-off is particularly problematic in clinical contexts where missing relevant MH indicators can have serious consequences, making the accuracy conditioned on the relevant class paramount despite the apparent improvement in overall accuracy.

Figure~\ref{multi}\textbf{(B)} illustrates how the logits of positive observations evolve as the steering strength increases from 0 to 2 (for visualization clarity, only positive steering strengths are shown to avoid overlapping trajectories), revealing how the model’s internal decision boundary shifts under different steering configurations.

\subsection{BDI-II Questionnaire Completion.}
To address \textbf{(RQ2)}, we evaluate the effectiveness of our steered LLM on the downstream task of automated BDI-II questionnaire completion. 
To do that we use the official eRisk benchmark metrics \cite{losada2019overview} that assess performance at two distinct levels (for all the metrics considered, the higher the value, the better):  

At \textit{the level of the questionnaire}, we use the Hit Rate of the Depression Category (DCHR) which measures the accuracy in estimating depression severity levels (minimal: \textbf{0-9}, mild: \textbf{10-18}, moderate: \textbf{19-29}, and severe: \textbf{30-63}, according to the eRisk competition), and the Average Difference between Overall Depression Levels (ADODL), which evaluates the general BDI-II score estimations. 
DCHR and ADODL represent the most clinically meaningful metrics, as they directly capture the model’s ability to assign users to the correct depression severity category and to approximate the total clinical score, both of which are essential for ensuring that automated assessments align with established diagnostic screening standards.

At \textit{item level}, we use Average Hit Rate (AHR) to evaluate prediction accuracy for individual symptoms, and Average Closeness Rate (ACR) to measure how close predictions are to actual values for each symptom.

Since the number of Reddit posts per user varies considerably (often exceeding the context window of standard LLMs) we adopt the adaptive Retrieval-Augmented Generation (\textit{aRAG}) framework proposed by Ravenda et al.~\cite{ravenda-etal-2025-llms}. 
This method performs item-specific adaptive retrieval, dynamically selecting the most semantically relevant posts for each BDI-II item rather than using a fixed retrieval size.

Given a set of user posts $\mathcal{P}_i = \{p_{i1}, \dots, p_{im}\}$ and the textual representation of an item $q_j$ (where $j = 1, ..., 21$ as the number of items in BDI-III), we compute similarities
\[
s(p_{ik}, q_j) = \cos(f(p_{ik}), f(q_j)),
\]
where $f(\cdot)$ is the dense retrieval model from which we extract the embeddings (in this case of this work, we use \\ \texttt{msmarco-MiniLM-L-12-v3}\footnote{\href{https://huggingface.co/sentence-transformers/msmarco-MiniLM-L-12-v3}{https://huggingface.co/sentence-transformers/msmarco-MiniLM-L-12-v3}}). The top-$k_{ij}^*$ posts of user $i$ are then retrieved adaptively, with $k_{ij}^*$ varying across items based on the local neighborhood structure of the query in the embedding space and the density of its nearest neighbors post representations.

\begin{table*}[t]
    \centering  
    \setlength{\tabcolsep}{4pt} 
    \begin{tabular}{llcccc}
        \toprule
        Collection & Model   &\multicolumn{2}{c}{Questionnaire Metrics} &  \multicolumn{2}{c}{Item Metrics} \\
        \cmidrule(lr){1-4} \cmidrule(lr){5-6}
        & &DCHR  & ADODL  &  AHR  & ACR \\
        \midrule
        \multirow{9}{*}{\rotatebox[origin=c]{90}{eRisk 2021}} 
        &DUTH (Spartalis et al. \cite{spartalis2021transfer})&  15.00\% & 73.97\%  & 35.36\% & 67.18\% \\
&Symanto (Basile et al. \cite{basile2021upv})&   32.50\% & 82.42\% & 34.17\% & \textbf{73.17}\% \\
& CYUT (Wu and Qiu \cite{wu2021roberta})  &   41.25\% & 83.59\%  & 32.62\% & 69.46\% \\
        \cdashline{2-6}[1pt/1pt]
        & \texttt{Llama 3.1 8B Steered}   &   48.75\% & \textbf{83.63}\%  & 34.23\%  & 71.01\%  \\
        & \texttt{qwen-2.5-72b } &   46.25\% & 82.91\%  & 34.76\%  & 70.10\%  \\
        & \texttt{gemini-2.5-flash-lite}  &   52.50\% & 83.55\%  & 32.74\%  & 67.94\%  \\
        &  \texttt{kimi-k2} &   \textbf{53.75}\% & 81.88\%  & \textbf{36.01}\%  & 70.85\%  \\
        & \texttt{mistral-medium-3.1} & 48.75\% & 80.24\%  & 34.76\%  & 70.00\%  \\

        \bottomrule
    \end{tabular}
        \caption{Model performance comparison on eRisk 2021 collection w.r.t. questionnaire metrics (DCHR, ADODL) and item metrics (AHR, ACR).  For all the metrics considered, higher scores indicate better performance. Bold values represent the best performance on specific metrics.}
    \label{model_comparison}
\end{table*}

The retrieved posts $\mathcal{P}_{i,j}^*$ are concatenated with the item text $q_j$ and passed to the LLM (steered according to the item-specific optimal strength $\lambda_j^*$ for each item) for zero-shot item scoring\footnote{For \texttt{Llama~3.1~8B}, item scores are derived from next-token prediction (NTP) probabilities, assigning the final score to the Likert option (0–3) with the highest predicted likelihood. This formulation enforces a constrained decoding setup, ensuring deterministic selection among predefined options and preventing undesired free-form generations such as refusals or explanatory text.}:
\[
y_{i,j} = \mathrm{LLM}_{\text{steered}}\big([\mathcal{P}_{i,j}^*; q_j], \lambda_j^*\big)
\]

where $ y_{i,j}$ 
 denotes the predicted score (e.g., the 0–3 Likert value) for the BDI-II item ( $q_j$ )
This procedure ensures that each prediction relies on the most contextually relevant evidence while keeping prompts within the model’s token limit, enabling fair, and training-free clinical assessment. The prompt used is shown in Figure \ref{multi}\textbf{(C)}.

In Figure \ref{multi}\textbf{(D)} we show how our steered \texttt{Llama 3.1 8B} model outperforms the unsteered baseline and the steered versions when $\lambda = \{1, 2\}$, w.r.t. the DCHR metric.

We conducted a comprehensive comparison with state-of-the-art LLMs, both open- and closed- source models\footnote{All the considered LLMs were used with the temperature parameter set to 0 to force deterministic outputs.
}. Table \ref{model_comparison} presents the performance comparison on the eRisk 2021 collection for automated BDI-II questionnaire completion.

Our steered \texttt{Llama 3.1 8B} model demonstrates competitive performance across all metrics, achieving 48.75\% DCHR and 83.63\% ADODL. Notably, the model outperforms several significantly larger and up-to-date architectures, including the 72B-parameter \texttt{qwen-2.5-}\\ \texttt{-72b} (46.25\% DCHR) and matches the performance of \texttt{mistral-}\\ \texttt{medium-3.1} (48.75\% DCHR), while maintaining comparable results to other large-scale models such as \texttt{gemini-2.5-flash-lite} (52.50\% DCHR), and \texttt{kimi-k2} (53.75\% DCHR). 
For completeness, we also show results from top-performing approaches from eRisk 2021 competition.

Notably, our compact model achieves the best ADODL results among all compared models and ranks among the top performers in ACR metrics, demonstrating that steering vectors can enable smaller models to compete effectively with larger architectures in specialized MH assessment tasks.

\section{Conclusions \& Future Works}

Our experimental results demonstrate that steering vectors can effectively address the proposed research questions.
First, we show that steering systematically reduces the \textit{cautious bias} observed in compact LLMs, leading to less skewed estimators in zero-shot relevance prediction tasks \textbf{(RQ1)}. By introducing directional adjustments in the representation space rather than modifying model weights, our approach enables controlled behavioral modulation that improves performance in sensitive MH contexts.

Second, the same mechanism generalizes successfully to downstream BDI-II completion tasks \textbf{(RQ2)}, where our steered LLM exhibit improved alignment with clinically validated ground-truth responses and achieve performance comparable to substantially larger models. This finding highlights the potential of steering-based interventions as an efficient alternative to fine-tuning, particularly for resource-constrained deployments where computational cost and data availability are limiting factors.

Finally, we propose our data-driven approach to estimate the optimal steering strength \textbf{(RQ3)}, avoiding ad-hoc calibration. By calibrating steering to achieve a controlled balance between correction and stability.

Overall, these findings suggest that steering can serve as a lightweight yet powerful mechanism for adapting compact LLMs to domain-specific inference tasks (in this case MH assessment) without retraining or architectural modification. 

Although the results are promising, future work should extend this line of research beyond MH contexts and tasks, in order to further validate and generalize the effectiveness of our steering approach.

\section{Limitations \& Ethical Considerations}

Despite the promising results, this study has some limitations that open directions for future research.  
First, our approach assumes that the linear steering direction in the embedding space faithfully captures the latent decision geometry of the LLM. This assumption, while empirically supported by our results, remains a simplification of the complex and highly non-linear behavior of transformer-based models. 

Second, the steering intervention operates in a static fashion, applying a fixed steering vector throughout inference for each BDI-II item. In realistic scenarios, user-generated content is highly heterogeneous, and adaptive or context-dependent steering, modulated dynamically according to post content or uncertainty, may further enhance interpretability and performance.  

Finally, our evaluation focuses on English-language Reddit data. Cultural and linguistic variations in how psychological distress is expressed could influence both retrieval quality and model sensitivity. Cross-lingual or cross-cultural validation is therefore required to assess the generalizability of steering-based MH assessment.

The proposed methodology for MH support and assessment, while novel, entails several ethical implications that must be carefully considered to ensure responsible use.

The primary concern is the potential misuse of LLMs as clinical diagnostic tools. Without adequate oversight and safeguards, such systems could reinforce biases or produce misleading inferences with harmful consequences. This approach should therefore be regarded strictly as a decision-support framework (intended to assist early screening or research applications) not as a replacement for qualified MH professionals.

\printbibliography

\appendix

\end{document}